\documentclass[lettersize,journal]{IEEEtran}
\usepackage{amsmath,amssymb,amsfonts}
\usepackage{algorithm}
\usepackage{algpseudocode}
\usepackage{array}
\usepackage[caption=false,font=normalsize,labelfont=sf,textfont=sf]{subfig}
\usepackage{textcomp}
\usepackage{stfloats}
\usepackage{url}
\usepackage{verbatim}
\usepackage{graphicx}
\usepackage{cite}
\usepackage{bm}
\usepackage{xcolor}
\usepackage{multirow}
\usepackage{cleveref}
\usepackage[normalem]{ulem}

\graphicspath{{Figure/}{./}}

\begin{document}

\title{A Pseudo Global Fusion Paradigm-Based Cross-View Network for LiDAR-Based Place Recognition}

\author{Jintao~Cheng, Jiehao~Luo, Xieyuanli~Chen, Jin~Wu, Rui~Fan, Xiaoyu~Tang$^{\ddag}$, and Wei~Zhang
\thanks{$^{\ddag}$ Corresponding authors}
\thanks{Jintao Cheng and Xiaoyu Tang are with the School of Electronics and Information Engineering, and Xingzhi College, South China Normal University, Foshan 528225, China.}
\thanks{Jiehao Luo is with the School of Data Science and Engineering, and Xingzhi College, South China Normal University, Shanwei 516600, China.}
\thanks{Xieyuanli Chen is with the College of Intelligence Science and Technology, National University of Defense Technology, Changsha, China.}
\thanks{Rui Fan is with the College of Electronics \& Information Engineering, Shanghai Research Institute for Intelligent Autonomous Systems, the State Key Laboratory of Intelligent Autonomous Systems, and Frontiers Science Center for Intelligent Autonomous Systems, Tongji University, Shanghai 201804, China.}
\thanks{Jintao Cheng, Jin Wu, and Wei Zhang are also with the Department of Electronic and Computer Engineering, Hong Kong University of Science and Technology, Hong Kong, China.}}

\markboth{Journal of \LaTeX\ Class Files,~Vol.~XX, No.~XX, XXXX}%
{Cheng \MakeLowercase{\textit{et al.}}: A Pseudo Global Fusion Paradigm-Based Cross-View Network for LiDAR-Based Place Recognition}

\maketitle

\begin{abstract}
LiDAR-based Place Recognition (LPR) remains a critical task in Embodied Artificial Intelligence (AI) and Autonomous Driving, primarily addressing localization challenges in GPS-denied environments and supporting loop closure detection. Existing approaches reduce place recognition to a Euclidean distance-based metric learning task, neglecting the feature space's intrinsic structures and intra-class variances. Such Euclidean-centric formulation inherently limits the model's capacity to capture nonlinear data distributions, leading to suboptimal performance in complex environments and temporal-varying scenarios. To address these challenges, we propose a novel cross-view network based on an innovative fusion paradigm. Our framework introduces a pseudo-global information guidance mechanism that coordinates multi-modal branches to perform feature learning within a unified semantic space. Concurrently, we propose a Manifold Adaptation and Pairwise Variance-Locality Learning Metric that constructs a Symmetric Positive Definite (SPD) matrix to compute Mahalanobis distance, superseding traditional Euclidean distance metrics. This geometric formulation enables the model to accurately characterize intrinsic data distributions and capture complex inter-class dependencies within the feature space. Experimental results demonstrate that the proposed algorithm achieves competitive performance, particularly excelling in complex environmental conditions.
\end{abstract}


\begin{figure}[t!]
    \centering
    \includegraphics[width=0.95\columnwidth]{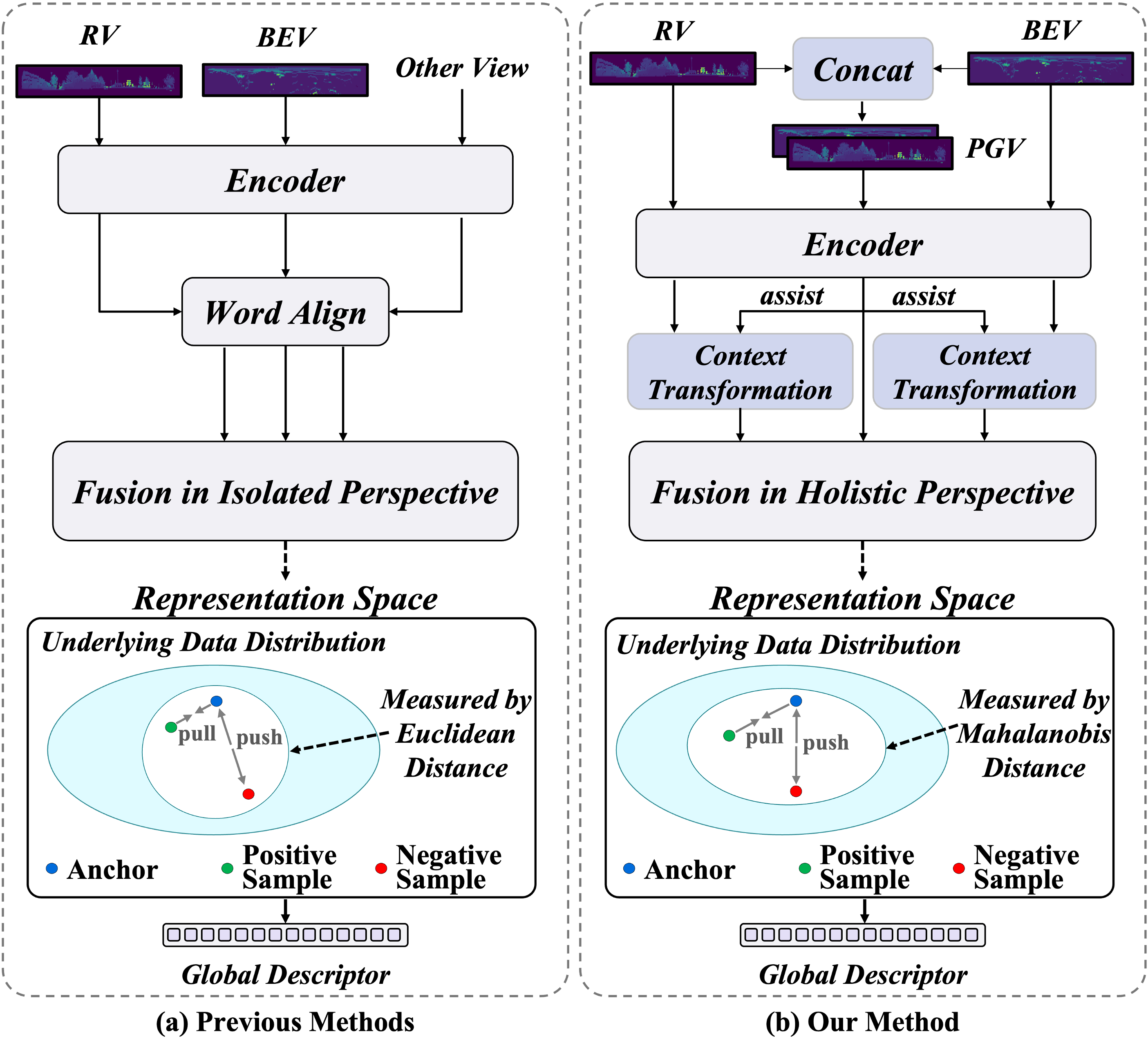}
    \caption{Core idea of the proposed pseudo-global fusion paradigm. The left side (a) illustrates previous methods with isolated feature extraction followed by late fusion and Euclidean distance metrics for optimization. The right side (b) shows our proposed method creates a pseudo-global representation, enabling early contextual information sharing and unified semantic learning. The proposed MAPVLM approach transforms the traditional Euclidean feature space into a more discriminative Mahalanobis space that better captures intrinsic geometric structure and intra-class variances, significantly improving place recognition performance in complex environments.}
    \label{fig1}
\end{figure}

\section{Introduction}
\label{sec:introduction}
\IEEEPARstart{T}{he} growing demand for autonomous navigation systems has positioned LiDAR-based place recognition (LPR)~\cite{chen_overlapnet_2020, scancontext, ma_seqot_2023} as a critical capability for long-term localization in Global Positioning System (GPS)-denied environments. While visual place recognition approaches struggle with illumination changes, LiDAR's geometric stability offers distinct advantages for cross-seasonal applications. LPR analyzes point cloud data to identify previously visited locations, serving two critical functions in autonomous systems: environmental perception and enhanced localization. This capability provides direct support for loop closure detection in Simultaneous Localization and Mapping (SLAM) systems, significantly improving positioning accuracy for robots and self-driving vehicles~\cite{mfmos}.

Traditional hand-crafted descriptor paradigms rely on manually designed geometric features and exhibit inherent limitations in adapting to complex urban dynamics. Therefore, mainstream approaches aim to propose learning-based algorithms to capture environmental invariants implicitly. Current mainstream approaches typically project LiDAR data into multiple distinct views, such as Bird's Eye View (BEV) and Range View (RV), to extract view-specific features, followed by cross-view feature fusion to enhance overall performance. For instance, Ma et al.~\cite{ma_cvtnet_2024} developed a multi-view fusion framework where synchronized BEV and RV initially capture point cloud features, which are then semantically integrated through a transformer-based cross-attention module. While these methods partially mitigate information loss inherent in single-view approaches, their decoupled feature extraction paradigm fundamentally restricts the model's capacity to learn and integrate global contextual information during initial feature extraction stages, creating an "information island" problem. In this paradigm, each view processes data independently without cross-view guidance, preventing the model from exploiting complementary geometric information that could enhance place discrimination accuracy.

Commonly, descriptor-based place recognition algorithms are fundamentally metric learning tasks, where both feature representation and distance metric play pivotal roles in determining localization performance. However, most existing methods oversimplify place recognition as a Euclidean distance-based metric learning task, failing to account for intrinsic structures and intra-class variances in feature space. This geometric simplification fundamentally limits their ability to capture nonlinear data distributions, particularly in complex real-world environments. Given this context, feature learning guided by global information has been extensively utilized in recent studies. Wang et al.~\cite{wang_prfusion_2024} demonstrated the advantages of feature learning in a unified semantic space. However, their method still relies on Euclidean distance metrics, failing to capture the intrinsic geometric structure of the data. In summary, the conventional Euclidean distance metric assumes a homogeneous feature space, neglecting correlations between different dimensions and their varying importance. This proves particularly inadequate when processing LiDAR data with complex spatio-temporal variations.

Given this, we propose a cross-view framework enabling global-context-aware feature learning via a novel fusion paradigm. As illustrated in Fig. \ref{fig1}, different from previous algorithms' fusion features in isolated perspective, our proposed method presents a novel Pseudo-Global View Construction paradigm to perform contextual alignment within a unified global semantic space, overcoming the limitations of conventional cross-view approaches. Additionally, to address the issue of traditional Euclidean distance-based metric learning, we introduce a Manifold Adaptation and Pairwise Variance-Locality Learning Metric (MAPVLM) that constructs a Symmetric Positive Definite (SPD) matrix to compute Mahalanobis distance, superseding traditional Euclidean metrics. This geometric modeling enables our proposed method to more accurately characterize the intrinsic data distribution and inter-class dependencies, demonstrating superior performance in complex environments and time-varying scenarios.

The main contributions of this work are threefold:
\begin{enumerate}
   \item We propose a novel fusion paradigm for LPR that introduces a pseudo-global view construction mechanism.
   \item We propose the Manifold Adaptation and Pairwise Variance-Locality Learning Metric, which constructs an adaptive SPD matrix for computing Mahalanobis distances.
   \item Experimental results demonstrate that our proposed method achieves competitive performance on multiple public datasets including NCLT, KITTI, and Ford Campus. The code has been open sourced.
\end{enumerate}

\section{Related Work}
\subsection{LiDAR-based Place Recognition}
LiDAR-based place recognition constitutes a pivotal technology in autonomous driving, robotic navigation, and spatial target localization~\cite{10032819, 9573287,ma_cvtnet_2024}. Early LPR methodologies predominantly relied on hand-crafted features and geometric descriptors. For example, Scan Context~\cite{scancontext} recorded the 3D structural configurations of sensor-visible spaces without requiring histograms or pretraining procedures. With the proliferation of deep learning techniques, neural network-based approaches have gradually emerged as the predominant paradigm. NetVLAD~\cite{arandjelovic_netvlad_2016} and PointNetVLAD~\cite{uy_pointnetvlad_2018} extract point cloud features and generate global descriptors through enhanced Vector of Locally Aggregated Descriptors (VLAD) layers and PointNet~\cite{qi2017pointnet} architectures, facilitating the automated learning of more discriminative feature representations, although with increased computational complexity.

To address the inherent limitations of point cloud-based methodologies in real-time data processing, researchers have explored projection-based approaches, transforming high-dimensional point cloud data into low-dimensional image representations. OverlapMamba~\cite{xiang_overlapmamba_2024} and CVTNet~\cite{ma_cvtnet_2024} generate heading-invariant global descriptors by fusing RV and BEV projections derived from LiDAR data. However, the 3D-to-2D transformation process potentially results in the loss of critical spatial information, which consequently affects recognition accuracy.

To mitigate the limitations inherent to single-source data, multi-modal fusion approaches have emerged as a prominent research direction. PRFusion~\cite{wang_prfusion_2024} innovates the data processing pipeline through multi-modal data fusion, and SLGD-Loop~\cite{arshad_slgd-loop_2024} addresses place recognition challenges in dynamic environments by integrating semantic information with deep learning. These methodologies demonstrate enhanced resilience to environmental variations and occlusion issues. Addressing challenges in large-scale and dynamic environments, Li et al.~\cite{li_hybrid_2018} proposed an approach that combines particle filtering and Kalman filtering techniques. SeqOT~\cite{ma_seqot_2023} and RING++~\cite{xu_ring_2023} demonstrate the efficacy of multi-scale feature fusion and learning-free frameworks in managing large-scale environments. MVSE-Net~\cite{zhang_mvse-net_2024} exhibits robust generalization capabilities across diverse datasets by combining multi-view projections with semantic embeddings.

Despite significant advancements in existing methodologies, most adopt view-level feature alignment and late fusion strategies, creating an "information island" problem. This isolation prevents each view from leveraging complementary geometric cues—such as BEV's spatial layout awareness and RV's range-height relationships—during the critical early feature extraction phase, ultimately limiting the model's ability to capture comprehensive environmental representations. These approaches fail to position different views within a unified global semantic space for interactive learning, thus inadequately expressing the complex structures of feature spaces and limiting their ability to learn and integrate global contextual information during early feature extraction stages.

\begin{figure*}[t!]
    \centering
    \includegraphics[width=0.95\textwidth]{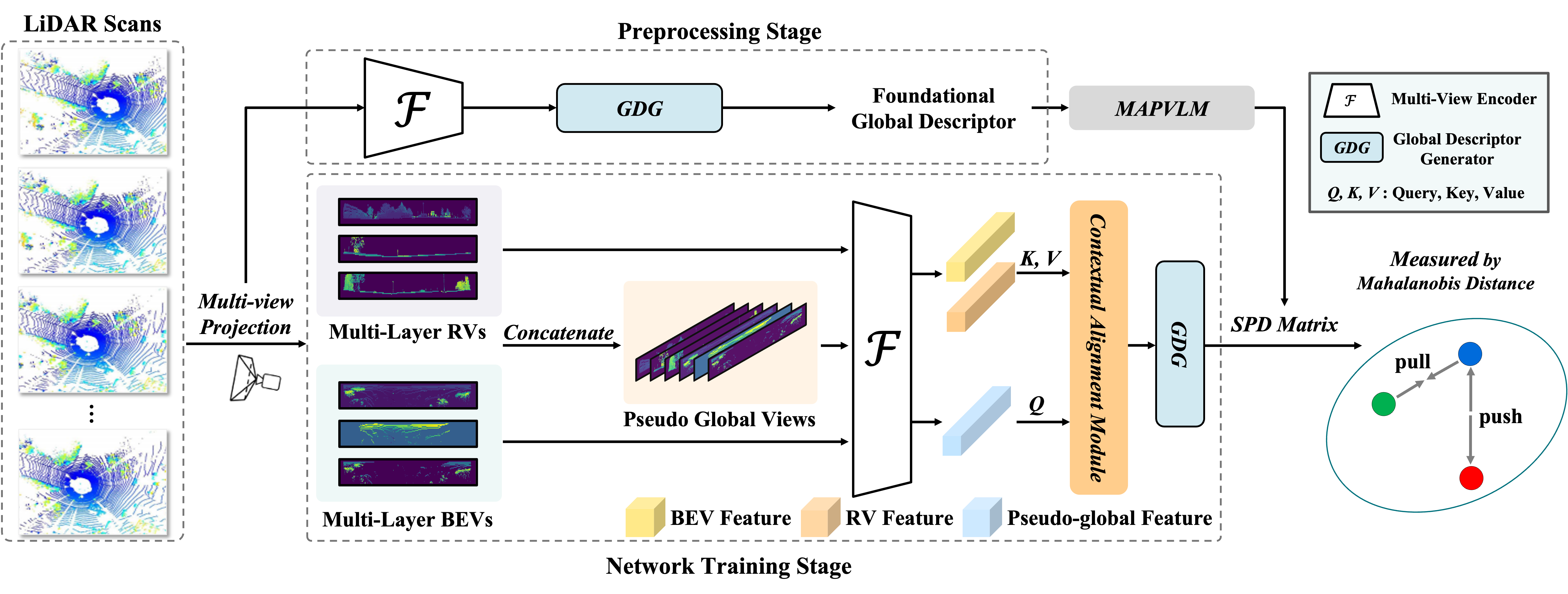}
    \caption{Framework architecture of our proposed method. The framework consists of three main components: preprocessing for global descriptor initialization, pseudo-global fusion for feature enhancement, and MAPVLM for learning an adaptive SPD matrix. Compared to conventional methods, our approach addresses both the information island problem through early global context sharing and the metric learning limitation through Mahalanobis distance supervision.}
    \label{fig2}
\end{figure*}

\subsection{SPD-assisted Metric Learning}
Metric learning represents a classical task in the domains of computer vision and machine learning~\cite{9976232, 9459530}. Mainstream LPR methodologies are predominantly based on Euclidean distance for supervision. For example, CVTNet~\cite{ma_cvtnet_2024} employ triplet loss with Euclidean distance to learn discriminative global descriptors. However, Euclidean metrics assume uniformity in feature spaces, neglecting internal structures of data distributions and intra-class variance differentials. To address metric learning limitations, Gao et al.~\cite{gao_learning_2019} proposed a methodology to characterize the underlying structures of visual features through SPD matrices. Similarly, Omara et al.~\cite{omara_novel_2021} introduced an ear recognition approach based on learning Mahalanobis distance metrics from SPD matrices, demonstrating the advantages of more flexible measurement approaches. Although these methodologies significantly enhance metric effectiveness, their application contexts remain constrained to the image recognition domain. In contrast, our research focuses on developing sample measurement methodologies for 1D descriptors applicable to LPR, overcoming the inherent limitations of traditional Euclidean metrics.

In SPD matrix-based metric learning, dimension reduction constitutes a necessary step to alleviate burdens of matrix computation. Local Fisher Discriminant Analysis (LFDA) effectively considers the local structural characteristics of data samples by combining the advantages of Fisher Discriminant Analysis (FDA) and Locality Preserving Projection (LPP). Zhong et al.~\cite{zhong_fault_2020} significantly improve fault diagnosis performance and model interpretability by exploiting local data structural features from both sample and variable dimensions. However, as a linear methodology, LFDA faces challenges in describing complex nonlinear systems. To overcome this limitation, kernel-based nonlinear LFDA (KLFDA)~\cite{van_bearing_2016} has garnered extensive attention. And multiple studies~\cite{DBLP:journals/corr/VemulapalliJ15} have applied Principal Component Analysis (PCA) as a preprocessing methodology for visual representations. Directly applying LFDA to high-dimensional features may result in inaccurate covariance matrix estimation or excessive computational complexity, whereas PCA preprocessing can eliminate redundant and noisy dimensions, providing a more stable and compact input feature space for LFDA, which provides an important theoretical foundation and developmental direction for our research.

\section{Methodology}
\subsection{Overview}
As illustrated in Fig. \ref{fig2}, the proposed framework could be divided into three parts: pre-processing stage, pseudo global fusion-based feature enhancement, and Mahalanobis distance supervision via SPD matrix assistant. Firstly, we introduce our pre-processing phase, which aims to obtain relatively accurate global descriptors for optimizing the main framework. Next, the proposed pseudo-fusion network is introduced to solve the lack of accurate global information. Next, to address the critical limitation of global context deficiency in existing approaches, we propose a novel pseudo-fusion network architecture. Simultaneously, the proposed MAPVLM, which constructs a SPD matrix to compute Mahalanobis distances, thereby superseding conventional Euclidean distance metrics.

\subsection{Pre-processing Stage}
We implement a pre-processing stage to establish foundational global descriptors for place recognition. Following a similar approach to CVTNet~\cite{ma_cvtnet_2024}, we project the original LiDAR scans to BEV and RV representations, then process these projections through our feature extractor (detailed in Section \ref{subsec:feature_extractor}) to obtain multi-dimensional feature embeddings. These extracted features are subsequently transformed into baseline global descriptors via our Global Descriptor Generator (GDG), providing the essential feature representations for our advanced metric learning methodology.

\subsection{Pseudo Global Fusion-based Feature Enhancement}
In this section, we introduce our pseudo-global fusion paradigm, which addresses the lack of global contextual guidance in the early feature extraction phase and insufficient feature interaction across contexts in conventional multi-view networks.

\subsubsection{Multi View Projection}
Following established approaches in~\cite{ma_cvtnet_2024}, we implement a multiview multilayer generator that captures information at different ranges and heights. This technique discretizes space into preset intervals, creating split spaces $\mathcal{R} = \{\mathcal{R}_1,...,\mathcal{R}_q\}$ for spherical coordinates (using range intervals $\{s_0,...,s_q\}$) and $\mathcal{B} = \{\mathcal{B}_1,...,\mathcal{B}_q\}$ for Euclidean coordinates (using height intervals $\{t_0,...,t_q\}$).

For a LiDAR point $p_k = (x_k, y_k, z_k)$, the projection to RV and BEV coordinates can be expressed as:
\begin{equation}
\label{eq:projection_rv}
\begin{pmatrix}
 u_k^r \\
 v_k^r
\end{pmatrix} = 
\begin{pmatrix}
 \frac{1}{2}[1 - \arctan(y_k, x_k)/\pi] w \\
 [1 - (\arcsin(z_k/r_k) + f_{up})/f] h
\end{pmatrix}
\end{equation}
\begin{equation}
\label{eq:projection_bev}
\begin{pmatrix}
 u_k^b \\
 v_k^b
\end{pmatrix} = 
\begin{pmatrix}
 \frac{1}{2}[1 - \arctan(y_k, x_k)/\pi] w \\
 [r'_k/f'] h
\end{pmatrix}
\end{equation}
where $r_k = \|p_k\|_2$ is the range measurement of the point, $f = f_{up} + f_{down}$ represents the vertical field-of-view of the sensor, $r'_k = \|(x_k, y_k)\|_2$ is the planar distance, $f'$ denotes the maximum sensing range, and $w$ and $h$ are the width and height of the resulting image, respectively.

In our implementation, we maintain equal dimensions for both projections ($u_k^r = u_k^b$), ensuring spatial alignment between corresponding columns across different view representations, which facilitates our subsequent fusion strategy.

\subsubsection{Pseudo-Global View Construction}
Conventional cross-view networks typically extract features from different views independently before fusing them at later stages, creating an "information island" problem where each branch lacks awareness of complementary information from other views. To efficiently introduce global contextual guidance early in the process, we construct a pseudo-global view (PGV) by concatenating multilayer BEV and RV representations, which can be defined as:
\begin{equation}
\label{eq:pgv}
G = [\mathcal{B}_1, \ldots, \mathcal{B}_q, \mathcal{R}_1, \ldots, \mathcal{R}_q]
\end{equation}
where $\mathcal{B}_j$ and $\mathcal{R}_i$ represent the BEV and RV projections corresponding to different split spaces, as defined in the previous section. This concatenated representation provides a holistic view that captures cross-modal information, enabling more unified semantic learning while maintaining higher computational efficiency compared to using raw point clouds.

\begin{figure}[t!]
    \centering
    \includegraphics[width=0.95\columnwidth]{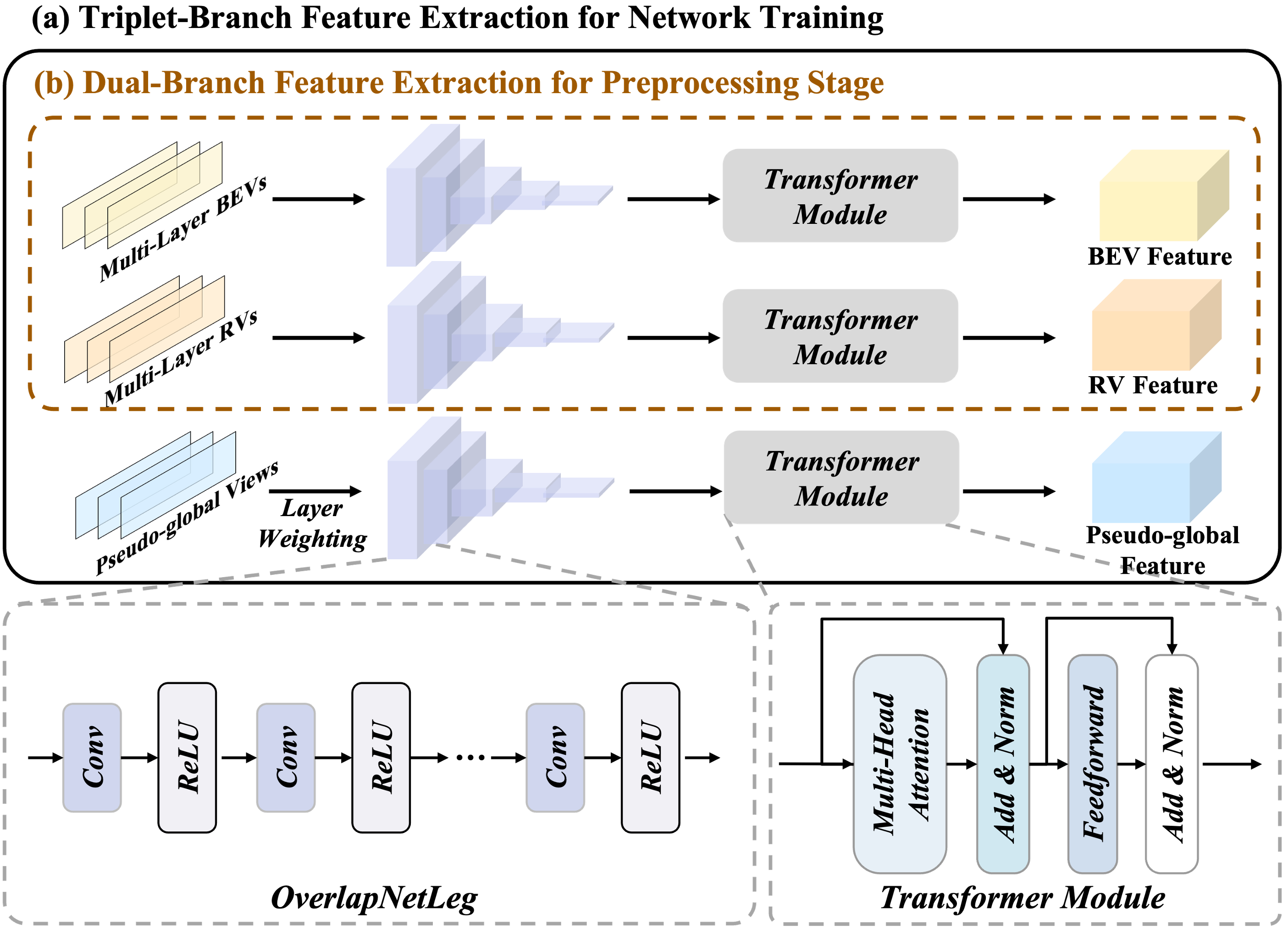}
    \caption{Feature extractor architecture. (a) The triple-branch configuration for network training incorporates the pseudo-global view alongside individual view branches. (b) The dual-branch structure is used during preprocessing. Both architectures utilize OverlapNetLeg for initial feature extraction, followed by Transformer modules for capturing long-range dependencies while maintaining spatial correspondence.}
    \label{fig:feature_extractor}
\end{figure}

\subsubsection{Feature Extractor}
\label{subsec:feature_extractor}
As shown in Fig. \ref{fig:feature_extractor}, we employ a dual-branch feature extractor for preprocessing and a triple-branch configuration for network training. Each branch consists of an OverlapNetLeg followed by a Transformer module. OverlapNetLeg \cite{ma_overlaptransformer_2022} uses convolutional layers to compress vertical dimensions while maintaining width, preserving spatial correspondence across different views. The Transformer module then captures long-range dependencies through self-attention mechanisms.

During preprocessing, the dual-branch structure processes BEVs and RVs independently, which can be represented as:
\begin{equation}
\label{eq:dual_branch}
\begin{aligned}
F_{BEV} &= \mathcal{T}_{BEV}(\mathcal{L}_{BEV}(\{\mathcal{B}_j\}_{j=1}^q)) \\
F_{RV} &= \mathcal{T}_{RV}(\mathcal{L}_{RV}(\{\mathcal{R}_i\}_{i=1}^q))
\end{aligned}
\end{equation}
where $\mathcal{L}$ represents the OverlapNetleg and $\mathcal{T}$ represents the Transformers module that processes features after initial extraction.

For network training, we add a third branch to process PGV. Different from \cite{ma_cvtnet_2024}, we introduce a simple layer weighting method to explicitly model projection channels with different distance or height thresholds. Given $\mathcal{G}$ with $2q$ layers, including $2q-2$ distance/height partitioned layers and 2 complete view, the layer weighting method and the process of the PGV can be expressed as:
\begin{gather*}
 w = \text{softmax}(\mathbf{W}\phi \cdot \text{GAP}(\mathcal{L}_{PGV}(G)) + \boldsymbol{\alpha}) \\
 \tilde{G} = \{\sum_{i=1}^{2q} w_i \cdot G_i\} + G
\end{gather*}
where each ${G}_i \in \mathbb{R}^{H \times W}$ represents a channel of $G$, $\text{GAP}(\cdot)$ denotes global average pooling, $\mathbf{W}_\phi \in \mathbb{R}^{2q \times C}$ is a learnable weight matrix, and $\boldsymbol{\alpha} \in \mathbb{R}^{2q}$ represents learnable layer importance biases. Finally, the PGV branch output is obtained by:
\begin{equation}
\label{eq:third_branch}
F_{PGV} = \mathcal{T}_{PGV}(\mathcal{E}_{PGV}(\tilde{G}))
\end{equation}
where $\mathcal{E}_{G}$ denotes the feature encoding module that processes PGV input while preserving its spatial structure and correspondence information.

Weights from pre-processing are transferred to the network training stage, accelerating convergence and facilitating early cross-view information sharing.

\begin{figure}[t!]
    \centering
    \includegraphics[width=\columnwidth]{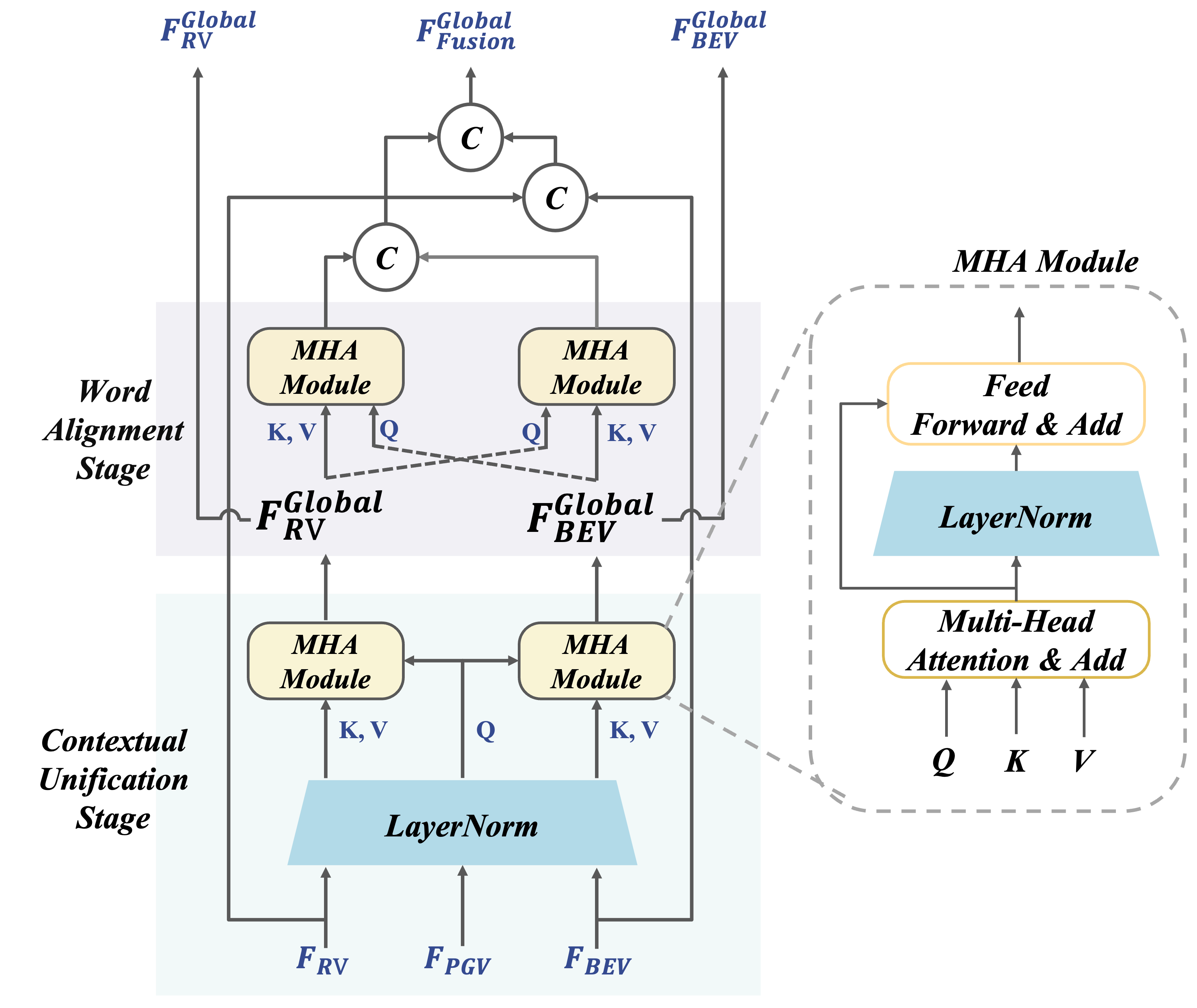}
    \caption{Contextual Alignment Module (CAM) architecture. The proposed CAM uses pseudo-global features as a query to guide the alignment of view-specific features through a two-stage process: Contextual Unification maps features into a global semantic space, and Word Alignment facilitates cross-view interactions under this unified context. This approach enables more effective integration of complementary information from different views compared to conventional late fusion strategies.}
    \label{fig5}
\end{figure}

\subsubsection{Contextual Alignment Module}
After obtaining features from both individual views and the PGV, we design a Contextual Alignment Module (CAM) based on cross-attention mechanisms, to effectively align these features within a unified semantic context while preserving their unique characteristics, as illustrated in Fig. \ref{fig5}.

The key insight behind our CAM design is that features extracted from PGV can act as a cross-modal mediator, guiding the alignment of $F_{BEV}$ and $F_{RV}$ through cross-attention mechanisms. This approach transforms features from their original modality-specific contexts into a unified global semantic space.

Given three common sets of inputs: query set $Q$, key set $K$, and value set $V$, we implement a standard multi-head attention mechanism $\mathcal{M}$, which can be expressed as:
\begin{equation}
\label{eq:mha}
\mathcal{M}(Q, K, V) = \text{softmax}\left(\frac{QK^T}{\sqrt{d_f}}\right)V
\end{equation}
where $d_f$ is a scaling factor.

Our CAM consists of two main stages: Contextual Unification and Word Alignment. In the Contextual Unification stage, we process the initial features through the multi-head attention modules, which can be represented as:
\begin{equation}
\label{eq:contextual_unification}
\begin{aligned}
F_{BEV}^{Global} &= \mathcal{M}(F_{PGV}, F_{BEV}, F_{BEV}) \\
F_{RV}^{Global} &= \mathcal{M}(F_{PGV}, F_{RV}, F_{RV})
\end{aligned}
\end{equation}
where $F_{BEV}^{Global}$ and $F_{RV}^{Global}$ represent the view-specific features mapped into the global semantic space.

By using the pseudo-global feature $F_{PGV}$ as Query and view-specific features as Key and Value pairs, we enable the pseudo-global context to selectively attend to relevant information in each view. This cross-attention mechanism effectively maps view-specific features into the global semantic space while also supplementing local views with complementary information from the global context.

In the subsequent Word Alignment stage, we further facilitate cross-view interactions under the unified global context, which can be defined as:
\begin{equation}
\label{eq:word_alignment}
\begin{aligned}
F_{BEV}^{Fusion} &= \mathcal{M}(F_{RV}^{Global}, F_{BEV}^{Global}, F_{BEV}^{Global}) \\
F_{RV}^{Fusion} &= \mathcal{M}(F_{BEV}^{Global}, F_{RV}^{Global}, F_{RV}^{Global}) \\
F_{Fusion}^{Global} &= {Concat}(F_{BEV}^{Fusion}, F_{PGV}, F_{RV}^{Fusion})
\end{aligned}
\end{equation}
The final fusion feature $F_{Fusion}^{Global}$ preserves the complementary information from both views while ensuring semantic consistency, which fundamentally differs from conventional late fusion strategies as it enables different view branches to learn features within a unified global context from the early stages, effectively addressing the information isolation problem in traditional cross-view networks.

\subsubsection{Global Descriptor Generator}
Following the Contextual Alignment Module, we employ a Global Descriptor Generator (GDG) to compress the fusion features into compact global descriptors. Similar to established approaches in LPR research~\cite{xiang_overlapmamba_2024, ma_cvtnet_2024}, our GDG consists of multiple parallel branches, each comprising a sequence of Multi-Layer Perceptron (MLP), NetVLAD, and MLP modules.

In our implementation, the components of the fused feature $F_{Fusion}^{Global}$ are processed through separate GDG branches. Each branch independently generates descriptors corresponding to its specific view representation. These branch-specific descriptors are then concatenated along the feature dimension to form the final global descriptor $D$. This branch-independent processing followed by concatenation strategy preserves the unique information from each view while the NetVLAD-MLP combination within each branch effectively captures statistical properties and reduces dimensionality.

\subsection{MAPVLM and Mahalanobis Distance Supervision}
\subsubsection{Manifold Adaptation and Pairwise Variance-Locality Learning Metric}
Conventional LPR methods relying on Euclidean distance metrics fail to capture intrinsic geometric structures of feature space and intra-class variance variations. We propose the MAPVLM that constructs an adaptive metric space by modeling nonlinear data distributions while preserving both pairwise variance differences and local neighborhood structures.

At the conclusion of the preprocessing stage, the network generates global descriptors $\{\mathbf{D}_i\}_{i=1}^N \in \mathbb{R}^{768}$ with corresponding location labels $\{y_i\}_{i=1}^N$. First, we compute the covariance matrix of the descriptor set, which can be defined as:
\begin{equation}
\label{eq:covariance}
\mathbf{C} = \frac{1}{N}\sum_{i=1}^{N}(\mathbf{D}_i - \bar{\mathbf{D}})(\mathbf{D}_i - \bar{\mathbf{D}})^T
\end{equation}
where $\bar{\mathbf{D}} = \frac{1}{N}\sum_{i=1}^{N}\mathbf{D}_i$ represents the mean descriptor across all samples in the dataset.

We then perform eigendecomposition on the covariance matrix to identify its principal components, which can be represented as:
\begin{equation}
\label{eq:eigen}
\mathbf{C} = \mathbf{U}\mathbf{\Lambda}\mathbf{U}^T
\end{equation}
where $\mathbf{U}$ contains the eigenvectors and $\mathbf{\Lambda}$ is a diagonal matrix of the corresponding eigenvalues.

To reduce dimensionality while preserving the most significant variance information, we select the top 256 eigenvectors to form the transformation matrix, which can be computed as:
\begin{equation}
\label{eq:transformation}
\mathbf{W}_1 = \mathbf{U}_{1:256}
\end{equation}

We apply this transformation to obtain lower-dimensional descriptor representations, which can be defined as:
\begin{equation}
\label{eq:reduced_dim}
\mathbf{D}_{reduced} = \mathbf{D}\mathbf{W}_1
\end{equation}
where $\mathbf{D}_{reduced} \in \mathbb{R}^{N \times 256}$ is the reduced-dimension representation of the original descriptors.

Next, we construct heat-kernel-based within-class and between-class affinity matrices to model local neighborhood relationships, which can be expressed as:
\begin{equation}
\label{eq:affinity_w}
\mathbf{A}_{ij}^{(w)} = 
\begin{cases}
\exp\left(-\frac{d_{ij}^2}{\sigma_{ij}^2}\right), & \text{if } y_i = y_j \\
0, & \text{otherwise}
\end{cases}
\end{equation}
\begin{equation}
\label{eq:affinity_b}
\mathbf{A}_{ij}^{(b)} = 
\begin{cases}
\exp\left(-\frac{d_{ij}^2}{\sigma_{ij}^2}\right), & \text{if } y_i \neq y_j \\
0, & \text{otherwise}
\end{cases}
\end{equation}
where $d_{ij} = \|\mathbf{D}_{reduced}^i - \mathbf{D}_{reduced}^j\|$ defines the Euclidean distance between the reduced-dimension descriptor vectors for samples $i$ and $j$, and $\sigma_{ij}$ is the adaptive bandwidth parameter defined as $\sigma_{ij} = \sigma_i \cdot \sigma_j$. The parameter $\sigma_i$ is computed as the distance between sample $i$ and its $k$-th nearest neighbor within the same class, adaptively adjusting the kernel bandwidth based on local data density.

Using these affinity matrices, we compute the weighted within-class and between-class scatter matrices, which can be represented as:
\begin{equation}
\label{eq:scatter_w}
\mathbf{S}_W = \frac{1}{2}\sum_{i,j=1}^{N} \mathbf{A}_{ij}^{(w)} d_{ij}d_{ij}^T
\end{equation}
\begin{equation}
\label{eq:scatter_b}
\mathbf{S}_B = \frac{1}{2}\sum_{i,j=1}^{N} \mathbf{A}_{ij}^{(b)} d_{ij}d_{ij}^T
\end{equation}
where $d_{ij} = \mathbf{D}_{reduced}^i - \mathbf{D}_{reduced}^j$.

These scatter matrices encode the variance structure within and between classes while preserving local neighborhood relationships.

To find the optimal discriminative projection, we solve the generalized eigenvalue problem, which can be formulated as:
\begin{equation}
\label{eq:generalized_eigen}
\mathbf{S}_B\mathbf{w} = \lambda\mathbf{S}_W\mathbf{w}
\end{equation}
where $\lambda$ represents the eigenvalues and $\mathbf{w}$ represents the corresponding eigenvectors.

We select the eigenvectors corresponding to the largest eigenvalues to form $\mathbf{W}_2 \in \mathbb{R}^{256 \times 256}$. Finally, we construct the Symmetric Positive Definite (SPD) matrix that encodes the learned metric space, which can be computed as:
\begin{equation}
\label{eq:spd_matrix}
\mathbf{M} = \mathbf{W}_1\mathbf{W}_2\mathbf{W}_2^T\mathbf{W}_1^T
\end{equation}

This SPD matrix $\mathbf{M}$ enables more accurate characterization of feature space relationships by capturing both the global variance structure through $\mathbf{W}_1$ and the local discriminative information through $\mathbf{W}_2$, all while maintaining computational efficiency.

\subsubsection{Mahalanobis Distance Metric}
The traditional Euclidean distance metrics constrain the model's ability to capture complex geometric structures in feature space. To overcome this issue, we leverage the SPD matrix $\mathbf{M}$ constructed during the preprocessing stage to define a more geometrically aware metric. The Mahalanobis distance between two feature vectors $\mathbf{f}_i$ and $\mathbf{f}_j$, which can be defined as:
\begin{equation}
\label{eq:mahalanobis}
 d_M(\mathbf{f}_i, \mathbf{f}_j) = \sqrt{(\mathbf{f}_i - \mathbf{f}_j)^T\mathbf{M}(\mathbf{f}_i - \mathbf{f}_j)}
\end{equation}
where $\mathbf{M}$ is the SPD matrix that encodes the learned metric space.

The diagonal elements of $\mathbf{M}$ weight feature dimensions according to their discriminative power, addressing Euclidean distance's uniform weighting limitation. Meanwhile, the off-diagonal elements capture correlations between dimensions, adapting to the data's intrinsic manifold structure.

\subsubsection{Loss Function}
Our training objective employs a triplet loss function utilizing the Mahalanobis distance, which can be expressed as:
\begin{align}
\label{eq:triplet_loss}
 L_T(G_q, \{G_p\}, \{G_n\}) = \alpha &+ \max_p(d_M(G_q, G_p)) \nonumber\\
 &- \min_n(d_M(G_q, G_n))
\end{align}
where $G_q$ represents the query descriptor, $\{G_p\}$ is the set of positive samples, $\{G_n\}$ is the set of negative samples, and $\alpha$ is a margin parameter that ensures a minimum separation between positive and negative pairs.

\section{Experiment}
\label{sec:experiments}
\subsection{Experimental Setup}
\subsubsection{Datasets}
We evaluate our proposed framework on three widely-adopted datasets representing diverse environments. The NCLT dataset~\cite{nclt} serves as an ideal testbed for long-term place recognition with temporal coverage across seasons and weather conditions. We utilize the 2012-01-08 sequence for training and 2012-02-05, 2012-06-15 , 2013-02-23 and 2013-04-05 sequences for evaluation. For loop closure detection, we employ the KITTI Odometry dataset~\cite{kitti} featuring urban environments captured by a 64-beam LiDAR, training on sequences 03-10, validating on sequence 02, and testing on sequence 00. We also access cross-environment generalization on Ford Campus dataset~\cite{pandey2011ford} sequence 00 without fine-tuning.

\begin{figure*}[t]
    \centering
    \includegraphics[width=0.9\textwidth]{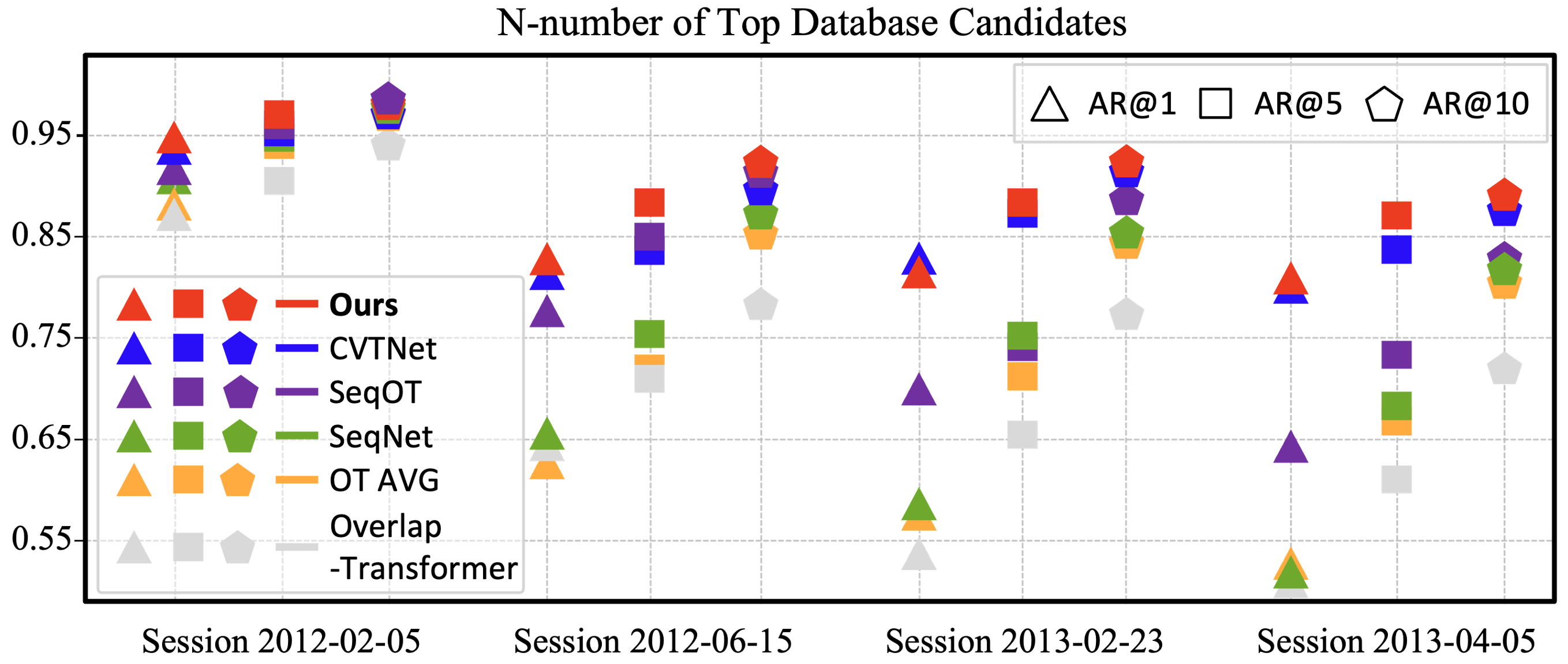}
    \caption{Performance comparison of place recognition task on the NCLT 2012-02-05, 2012-06-15, 2013-02-23, 2013-04-05 sequence.}
    \label{fig:ar_comparison_nclt}
\end{figure*}

\subsubsection{Evaluation Metrics}
For place recognition on NCLT, we adopt Average Recall@N (AR@N) for N={1,5,20}, indicating the probability of finding correct matches within top N candidates. For loop closure detection on KITTI, we employ standard metrics including Area Under Curve (AUC), maximum F1 score (F1max), and Recall@1/1\% to thoroughly evaluate retrieval performance.

\subsubsection{Implementation Details}
Our multi-view preprocessing utilizes a classic cross-view network~\cite{ma_cvtnet_2024}. For RV projection, we generated multi-layer depth images with dataset-specific intervals: NCLT uses [0-15m], [15-30m], [30-45m], and [45-60m], while KITTI and Ford Campus extend to 80m. For BEV projection, we defined height intervals: NCLT uses [-4-0m], [0-4m], [4-8m], and [8-12m], while KITTI and Ford Campus use [-3--1.5m], [-1.5-0m], [0-1.5m], and [1.5-5m]. Our network uses OverlapNetLeg~\cite{chen_overlapnet_2020} with transformer architecture as the feature extractor. For Mahalanobis distance computation, we constructed the SPD matrix using training sequences of each dataset.

\subsection{Place Recognition Performance}
The experimental results demonstrate that our proposed framework achieves superior performance across all evaluation metrics and temporal scenarios. As illustrated in Fig. \ref{fig:ar_comparison_nclt}, our proposed method exhibits consistent improvements across all four evaluation sequences.

\subsection{Loop Closure Detection Performance}
\begin{table*}[t]
\caption{Comparison of loop closure detection performance}
\label{tab:loop_closure}
\centering
\renewcommand{\arraystretch}{1.2}
\begin{tabular}{lcccccccc}
\hline
\multirow{2}{*}{Method} & \multicolumn{4}{c}{KITTI} & \multicolumn{4}{c}{Ford Campus} \\
\cline{2-9}
& AUC & F1max & Recall@1 & Recall@1\% & AUC & F1max & Recall@1 & Recall@1\% \\
\hline
Histogram~\cite{histogram} & 82.4 & 82.3 & 73.6 & 86.9 & 84.3 & 80.2 & 81.0 & 89.5 \\
Scan Context~\cite{scancontext} & 83.4 & 83.7 & 82.2 & 86.7 & 90.1 & 84.0 & 87.6 & 95.6 \\
LiDAR Iris~\cite{lidar_iris} & 84.1 & 84.6 & 83.3 & 87.5 & 90.5 & 84.4 & 84.7 & 93.5 \\
PointNetVLAD~\cite{uy_pointnetvlad_2018} & 85.8 & 84.4 & 77.8 & 84.7 & 87.4 & 82.8 & 86.0 & 93.6 \\
OverlapNet~\cite{chen_overlapnet_2020} & 86.5 & 86.3 & 81.4 & 90.6 & 85.2 & 84.5 & 85.5 & 93.0 \\
NDT-Transformer~\cite{ndttransformer} & 85.3 & 85.1 & 80.4 & 87.1 & 83.7 & 84.8 & 89.8 & 92.5 \\
MinkLoc3D~\cite{minkloc3d} & 89.2 & 86.7 & 87.4 & \underline{91.8} & 87.3 & 85.3 & 87.6 & 94.0 \\
CVTNet~\cite{ma_cvtnet_2024} & \underline{90.7} & \underline{87.7} & \underline{90.6} & \underline{96.4} & \underline{92.3} & \underline{85.6} & \underline{91.4} & \underline{95.4} \\
\textbf{Ours} & \textbf{91.9} & \textbf{89.1} & \textbf{91.3} & \textbf{96.8} & \textbf{92.6} & \textbf{87.2} & \textbf{92.0} & \textbf{96.3} \\
\hline
\end{tabular}
\end{table*}

\subsection{Ablation Study}
To systematically validate the effectiveness of each proposed component, we conduct comprehensive ablation experiments focusing on two key components: Contextual Alignment Module (CAM) and MAPVLM.

\begin{table}[t]
\caption{Ablation study of different modules}
\label{tab:ablation}
\centering
\renewcommand{\arraystretch}{1.2}
\setlength{\tabcolsep}{6pt}
\begin{tabular}{c cc ccc}
\hline
\multirow{2}{*}{Methods} & \multicolumn{2}{c}{Component} & \multicolumn{3}{c}{Performance} \\
\cline{2-6}
 & CAM & MARVPL & AR@1 & AR@5 & AR@20 \\
\hline
Baseline & $-$ & $-$ & 0.934 & 0.949 & 0.958 \\
\hline
\multirow{2}{*}{\centering (i)} & \checkmark & $-$ & 0.937 & 0.951 & 0.961 \\
 & $-$ & \checkmark & 0.937 & 0.950 & 0.960 \\
\hline
\centering (ii) & \checkmark & \checkmark & \textbf{0.939} & \textbf{0.953} & \textbf{0.963} \\
\hline
\end{tabular}
\end{table}

\begin{figure}[t!]
    \centering
    \includegraphics[width=0.95\columnwidth]{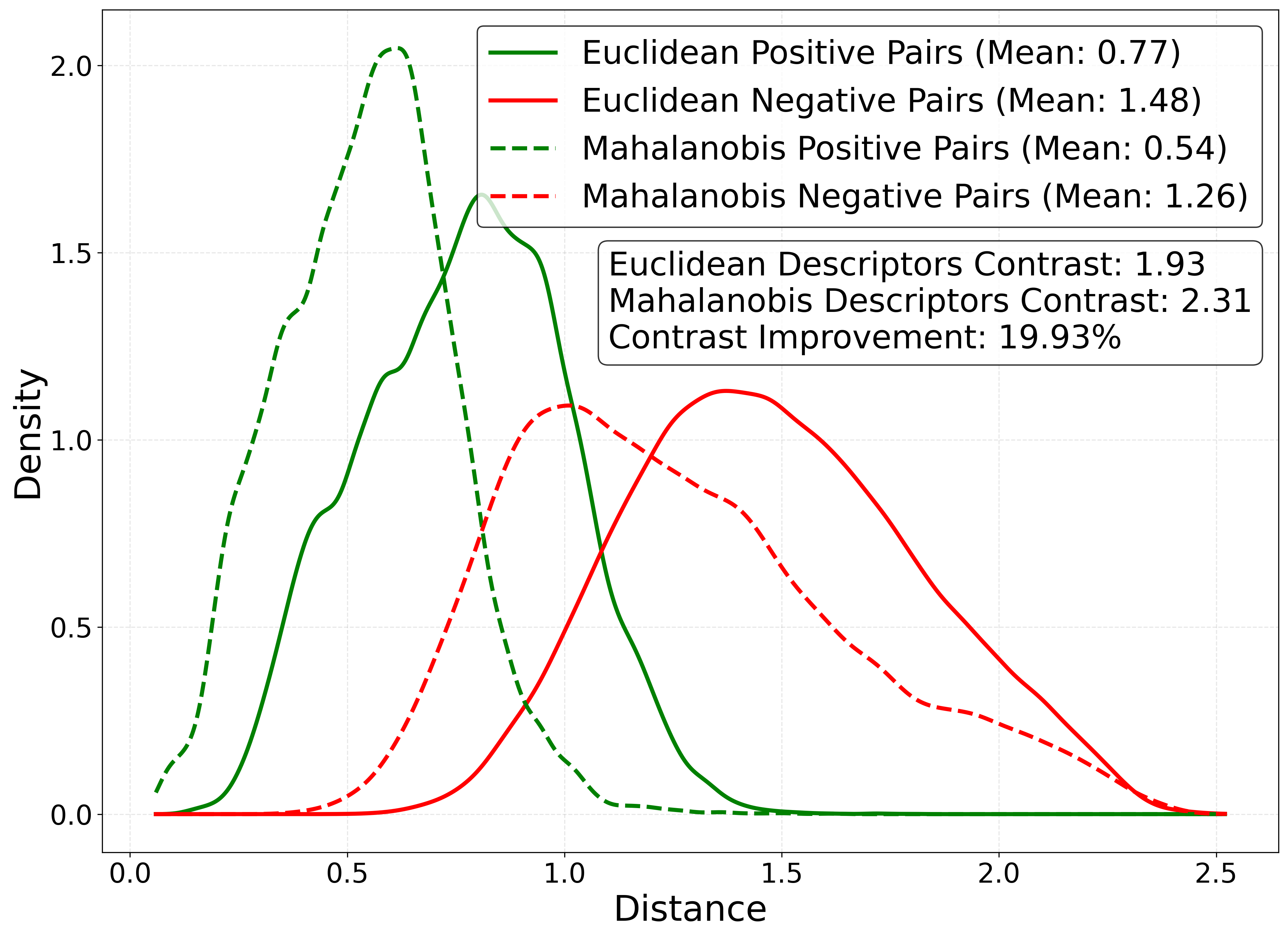}
    \caption{Distance distribution comparison between Euclidean and Mahalanobis metrics.}
    \label{fig:distribution}
\end{figure}

\begin{table}[t]
\caption{Performance comparison with and without MAPVLM}
\label{tab:mapvlm}
\centering
\renewcommand{\arraystretch}{1.2}
\setlength{\tabcolsep}{12pt}
\begin{tabular}{l c c}
\hline
Method & Baseline & w/ MAPVLM \\
\hline
OverlapTransformer \cite{ma_overlaptransformer_2022} & 87.5 & 89.1 \\
SeqOT \cite{ma_seqot_2023} & 91.6 & 92.4 \\
OverlapMamba \cite{xiang_overlapmamba_2024}& 93.3 & 93.4 \\
CVTNet \cite{ma_cvtnet_2024} & 93.2 & 93.5 \\
\hline
\end{tabular}
\end{table}

\subsection{Runtime Analysis}
\begin{table}[t]
\caption{Runtime comparison (ms)}
\label{tab:runtime}
\centering
\renewcommand{\arraystretch}{1.2}
\begin{tabular}{l@{\hspace{8pt}}c@{\hspace{8pt}}c@{\hspace{8pt}}c}
\hline
Method & Descriptor Generation & Retrieval & Total Time \\
\hline
Scan Context~\cite{scancontext} & 54.36 & 462.78 & 517.14 \\
LiDAR Iris~\cite{lidar_iris} & 6.82 & 8763.41 & 8770.23 \\
PointNetVLAD~\cite{uy_pointnetvlad_2018} & 12.95 & 1.33 & 14.28 \\
OverlapNet~\cite{chen_overlapnet_2020} & 4.52 & 3015.92 & 3020.44 \\
NDT-Transformer~\cite{ndttransformer} & 14.82 & 0.46 & 15.28 \\
MinkLoc3D~\cite{minkloc3d} & 14.92 & 7.55 & 22.47 \\
CVTNet~\cite{ma_cvtnet_2024} & 14.12 & 17.31 & 31.43 \\
\textbf{Ours} & 14.62 & 16.07 & 30.54 \\
\hline
\end{tabular}
\end{table}

\section{Conclusion}
In this paper, we proposed a novel pseudo-global fusion paradigm and MAPVLM approach for LiDAR-based place recognition that effectively addresses the information island problem and Euclidean distance limitations. Our proposed method achieves competitive performance on NCLT, KITTI, and Ford Campus datasets while maintaining real-time processing capabilities. The strong cross-environment generalization without fine-tuning demonstrates our approach's practical value for autonomous systems.

\normalem
\bibliographystyle{IEEEtran}
\bibliography{IEEE-Journal/manuscript_ieee}

\begin{IEEEbiography}[{\includegraphics[width=1in,height=1.25in,clip,keepaspectratio]{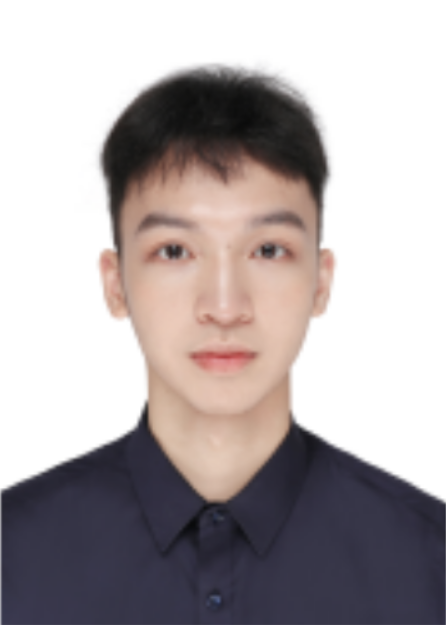}}]{Jintao Cheng}
received his bachelor's degree from the School of Physics and Telecommunications Engineering, South China Normal University, in 2021. He is currently pursuing an MPhil degree at The Hong Kong University of Science and Technology. His research includes computer vision, SLAM, and deep learning.
\end{IEEEbiography}

\begin{IEEEbiography}[{\includegraphics[width=1in,height=1.25in,clip,keepaspectratio]{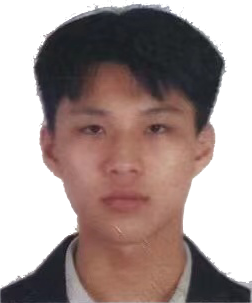}}]{Jiehao Luo}
(Student Member, IEEE) is currently conducting research under the supervision of Xiaoyu Tang at the School of Data Science and Engineering, and Xingzhi College, South China Normal University. His research focuses on computer vision and robotic perception.
\end{IEEEbiography}

\begin{IEEEbiography}[{\includegraphics[width=1in,height=1.25in,clip,keepaspectratio]{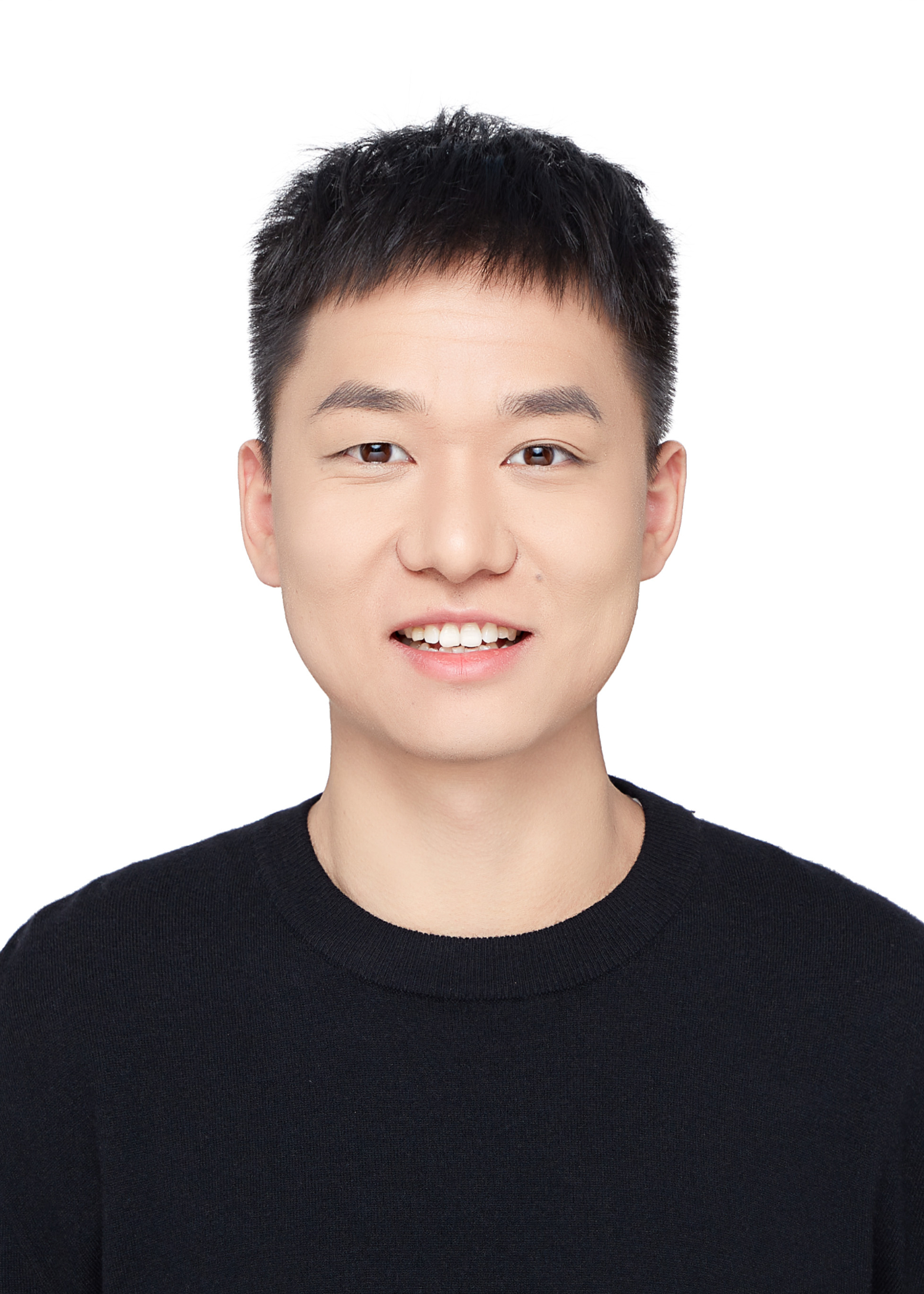}}]{Xieyuanli Chen}
Xieyuanli Chen is now an associate professor at the National University of Defense Technology, China. He received his Ph.D. degree in robotics at the Photogrammetry and Robotics Laboratory, University of Bonn. He received his master's degree in robotics in 2017 at the National University of Defense Technology, China, and his bachelor's degree in electrical engineering and automation in 2015 at Hunan University, China. He currently serves as an associate editor for IEEE RA-L, ICRA and IROS.
\end{IEEEbiography}

\begin{IEEEbiography}[{\includegraphics[width=1in,height=1.25in,clip]{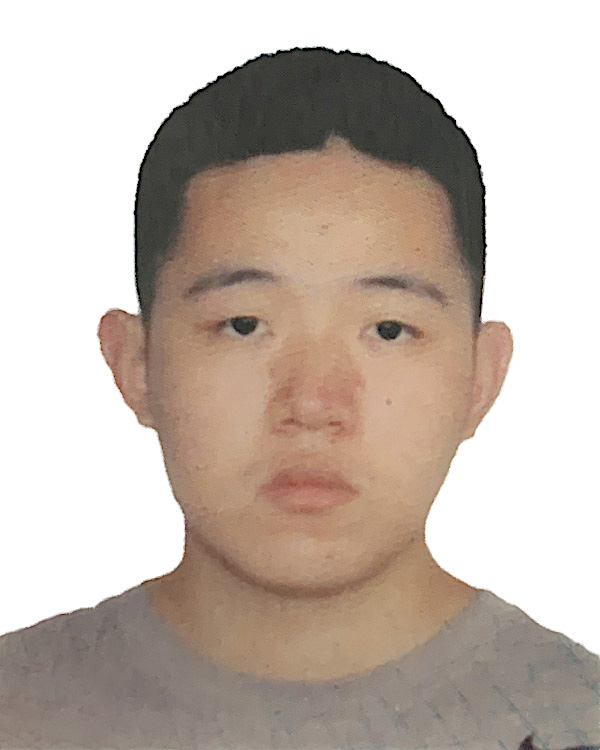}}]{Jin Wu}
(Member, IEEE) received a B.S. degree from the University of Electronic Science and Technology of China, Chengdu, China. From 2013 to 2014, he was a visiting student with Groep T, Katholieke Universiteit Leuven (KU Leuven). He is currently pursuing a Ph.D. degree in the Department of Electronic and Computer Engineering, Hong Kong University of Science and Technology (HKUST), Hong Kong, supervised by Prof. Wei Zhang. He has coauthored over 140 technical papers in representative journals and conference proceedings. He was awarded the outstanding reviewer of IEEE Transactions on Instrumentation and Measurement in 2021. He is now a review editor of Frontiers in Aerospace Engineering and an invited guest editor for several JCR-indexed journals. He is also an IEEE Consumer Technology Society (CTSoc) member and a committee member and publication liaison. He was a committee member for the IEEE CoDIT in 2019, a special section chair for the IEEE ICGNC in 2021, a special session chair for the 2023 IEEE ITSC, a track chair for the 2024 IEEE ICCE, 2024 IEEE ICCE-TW and a chair for the 2024 IEEE GEM conferences. From 2012 to 2018, he was in the micro air vehicle industry and has started two companies. From 2019 to 2020, he was with Tencent Robotics X, Shenzhen, China. He was selected as the World's Top 2\% Scientist by Stanford University and Elsevier in 2020, 2021 and 2022.
\end{IEEEbiography}

\begin{IEEEbiography}[{\includegraphics[width=1in,clip]{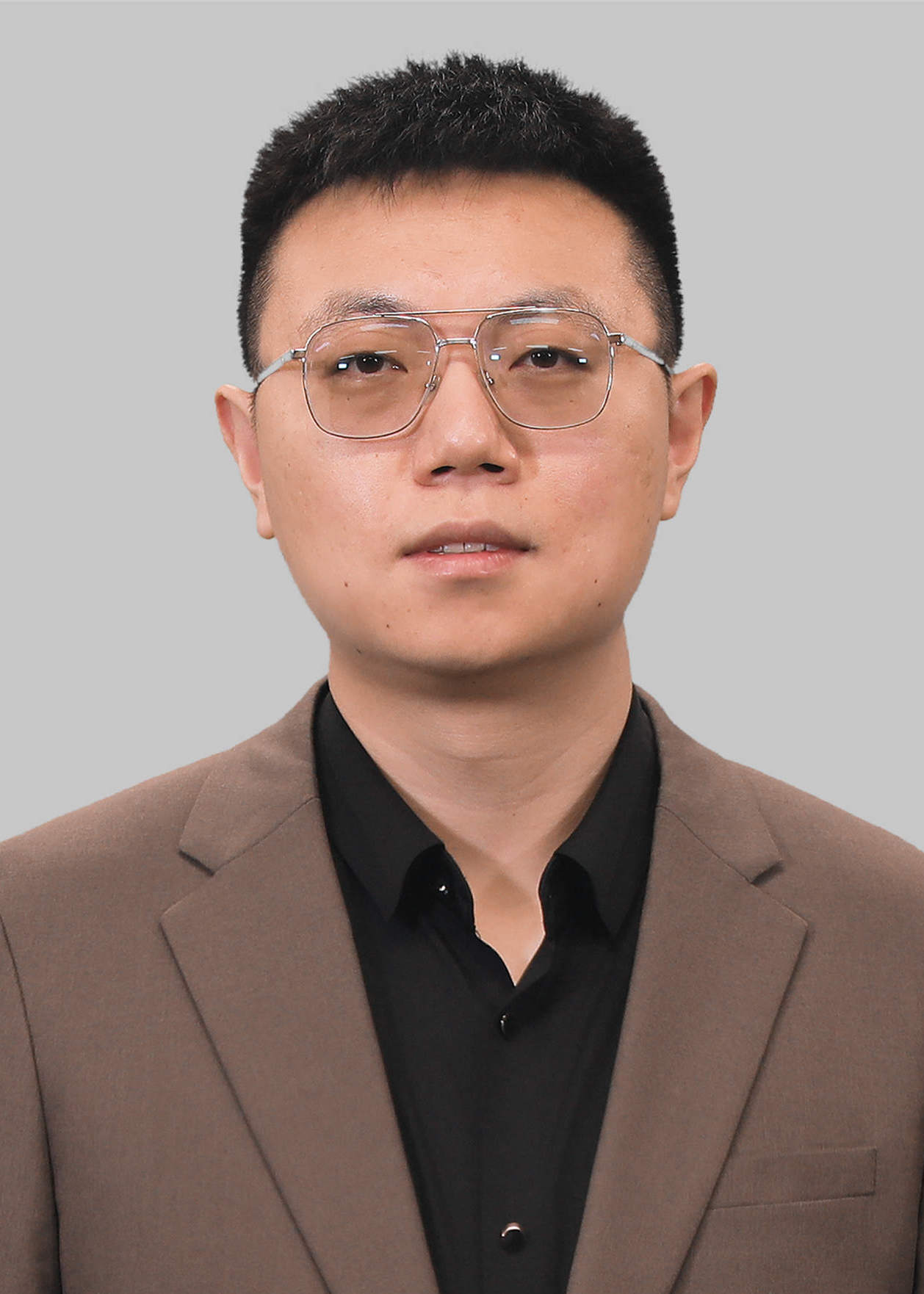}}]{Rui Fan}
(Senior Member, IEEE) received a B.Eng. degree in automation from the Harbin Institute of Technology in 2015 and a Ph.D. degree (supervisors: Prof. John G. Rarity and Prof. Naim Dahnoun) in Electrical and Electronic Engineering from the University of Bristol in 2018. He worked as a research associate (supervisor: Prof. Ming Liu) at the Hong Kong University of Science and Technology from 2018 to 2020 and a postdoctoral scholar-employee (supervisors: Prof. Linda M. Zangwill and Prof. David J. Kriegman) at the University of California San Diego between 2020 and 2021. He began his faculty career as a full research professor with the College of Electronics \& Information Engineering at Tongji University in 2021 and was then promoted to a full professor in the same college and at the Shanghai Research Institute for Intelligent Autonomous Systems in 2022.
\end{IEEEbiography}

\begin{IEEEbiography}[{\includegraphics[width=1in,height=1.25in,clip,keepaspectratio]{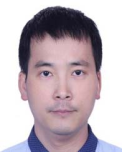}}]{Xiaoyu Tang}
(Member, IEEE) received a B.S. degree from South China Normal University, Guangzhou, China, in 2003 and an M.S. degree from Sun Yat-sen University, Guangzhou, China, in 2011. He is currently pursuing a Ph.D. degree with South China Normal University. Mr. Tang works as an associate professor, master supervisor, and deputy dean at Xingzhi College, South China Normal University. His research interests include image processing and intelligent control, artificial intelligence, the Internet of Things, and educational informatization.
\end{IEEEbiography}

\begin{IEEEbiography}[{\includegraphics[width=1in,height=1.25in,clip,keepaspectratio]{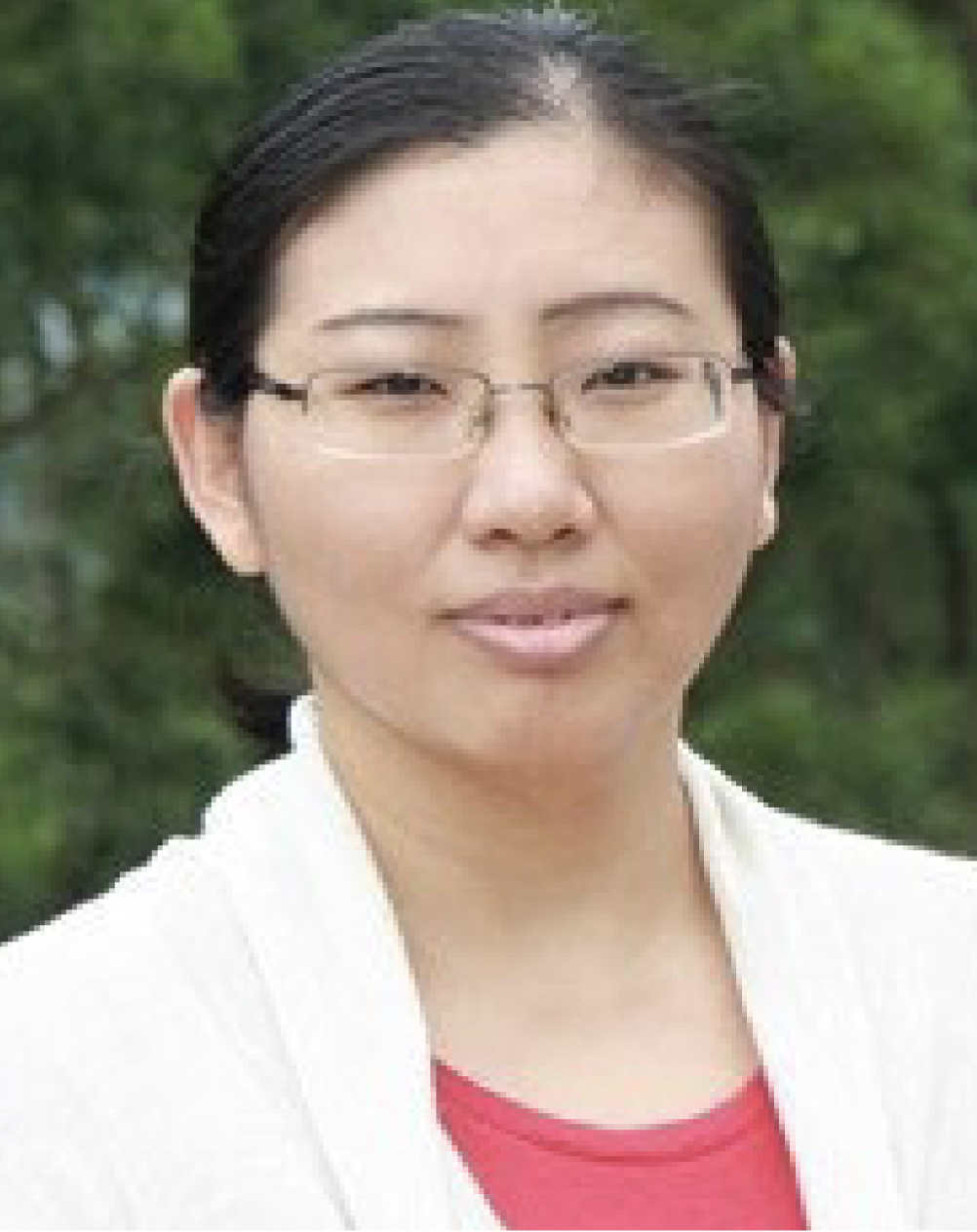}}]{Wei Zhang}
(Senior Member, IEEE) received the Ph.D. degree from Princeton University, Princeton, NJ, USA, in 2009. She is currently a Professor with the Department of Electronic and Computer Engineering, The Hong Kong University of Science and Technology, Hong Kong. She has authored over 150 book chapters and papers in peer reviewed journals and international conferences. Her current research interests include reconfigurable systems, computer architecture, EDA, and embedded system security.
\end{IEEEbiography}

\end{document}